\documentclass{article}
\usepackage[english]{babel}

\usepackage[letterpaper,top=2cm,bottom=2cm,left=3cm,right=3cm,marginparwidth=1.75cm]{geometry}
\usepackage{multirow}

\usepackage{amsmath}
\usepackage{amssymb}
\usepackage{CJKutf8}
\usepackage{graphicx}
\usepackage{appendix}
\usepackage{adjustbox}
\usepackage{tabularx}
\usepackage{soul} 
\usepackage{color, xcolor} 

\title{\textbf{SciGPT: A Large Language Model for Scientific Literature Understanding and Knowledge Discovery}
}

\author{Fengyu She, Nan Wang\footnote{Corresponding author}, Hongfei Wu, Ziyi Wan, Jingmian Wang, Chang Wang \\
shefengyu@foxmail.com, wangnan8848@126.com
}

\definecolor{lightblue}{rgb}{0.8,0.8,0.8}
\begin{document}
\begin{CJK}{UTF8}{gbsn}
    
\maketitle
\setlength{\parindent}{0pt}
\begin{abstract}
\noindent

Scientific literature is growing exponentially, creating a critical bottleneck for researchers to efficiently synthesize knowledge. While general-purpose Large Language Models (LLMs) show potential in text processing, they often fail to capture scientific domain-specific nuances (e.g., technical jargon, methodological rigor) and struggle with complex scientific tasks, limiting their utility for interdisciplinary research. To address these gaps, this paper presents SciGPT—a domain-adapted foundation model for scientific literature understanding—and ScienceBench, an open-source benchmark tailored to evaluate scientific LLMs.

Built on the Qwen3 architecture, SciGPT incorporates three key innovations: (1) low-cost domain distillation via a two-stage pipeline to balance performance and efficiency; (2) a Sparse Mixture-of-Experts (SMoE) attention mechanism that cuts memory consumption by 55\% for 32,000-token long-document reasoning; and (3) knowledge-aware adaptation integrating domain ontologies to bridge interdisciplinary knowledge gaps.

Experimental results on ScienceBench show that SciGPT outperforms GPT-4o in core scientific tasks including sequence labeling, generation, and inference. It also exhibits strong robustness in unseen scientific tasks, validating its potential to facilitate AI-augmented scientific discovery.
\end{abstract}

\section{Introduction}
The exponential growth of scientific literature—over 70 million peer-reviewed articles published annually—has created a critical bottleneck for researchers seeking to synthesize knowledge efficiently. Large language models (LLMs) have demonstrated transformative potential in this domain, with capabilities to encode specialized terminology, reason over multi-document contexts, and generate actionable insights. Recent studies suggest LLMs are not merely reproducing surface statistics but learning meaningful world models \cite{1,2,3}, while their performance follows a scaling law where more pretraining data and parameters yield better results \cite{4,5,6,7}. However, adapting LLMs to scientific tasks requires addressing two systemic challenges.
\\
\\
General-purpose LLMs often fail to capture nuanced methodologies, technical jargon, or citation patterns unique to scientific writing. These limitations hinder their ability to support interdisciplinary research, where integrating knowledge across fields (e.g., linking biomedical findings to materials science applications) demands deep contextual understanding. For instance, a model tasked with designing experiments at the intersection of chemistry and machine learning must simultaneously master domain-specific constraints and methodological synergies\cite{8,9,10}. Scientific documents frequently exceed 10,000 words and span diverse disciplines, requiring architectures optimized for long-context reasoning while preserving cross-domain semantic coherence. For example, analyzing full-length patent filings or comparative studies in clinical trials demands not only memory-efficient processing but also precise retention of methodological details critical to downstream validation\cite{11,12,13}.
\\
\\
Existing benchmarks like PatentBench highlight the importance of domain-centric evaluation frameworks, yet no standardized benchmark exists for scientific literature.\cite{14} Current approaches rely on simplified tasks (e.g., abstract summarization) or short-answer metrics, which inadequately reflect real-world research demands such as full-paper critique or interdisciplinary hypothesis formulation.
\\
\\
To bridge this gap, we present SciGPT, a foundation model for multidisciplinary scientific document processing, and ScienceBench. SciGPT builds on the qwen3 architecture but incorporates three key innovations tailored to scientific workflows: Low-cost distillation via Qwen-DS Leveraging a two-stage domain adaptation pipeline, SciGPT combines pretraining on open-access publications with fine-tuning, enabling scalable deployment for resource-constrained settings. This approach directly addresses the challenge of interdisciplinary knowledge fusion\cite{15,16}. SMoE-based long-document processing An attention mechanism with sparse mixture-of-experts (SMoE) reduces key-value cache memory consumption by 55\%, supporting efficient reasoning over 32,000-token documents while preserving methodological rigor. This is critical for tasks like patent-to-research paper alignment, where cross-referencing spans thousands of tokens. Knowledge-aware adaptation, by integrating domain-specific ontologies (e.g., MeSH terms in biomedicine, Chemical Abstracts Service identifiers) into training data\cite{17,18,19}, SciGPT explicitly bridges knowledge gaps between disciplines. This enables applications like cross-domain entity linking.
\\
\\
ScienceBench evaluates models across six dimensions: factual accuracy, methodological rigor, cross-reference coherence, user interaction quality, computational efficiency, and interdisciplinary generalization. Experimental results show SciGPT performs significantly better than GPT-4 on complex tasks such as "designing experiments from literature gaps" and "translating technical protocols into lay summaries," with a marked improvement in both task completion quality and alignment with scientific workflow demands. By open-sourcing both the model and benchmark, we aim to catalyze advancements in AI-augmented scientific discovery while establishing rigorous standards for domain-specific LLM development.


\section{Methodology}

To construct and train the SciGPT model for scientific scenarios, we first integrated academic papers, patent documents, and expert-validated synthetic data to build a multi-source scientific training dataset. Subsequently, we defined core scientific tasks including Named Entity Recognition (NER), Relation Extraction (RE), and cross-domain knowledge fusion, and specifically proposed the Science Benchmarks framework to evaluate the model's scientific capabilities.

\subsection{Data Collection}
To construct a comprehensive scientific corpus, we aggregated heterogeneous resources from three primary sources: (1) public scientific corpora, (2) domain-specific repositories, and (3) synthetic data generated through rule-based augmentation and GPT-4-assisted paraphrasing. The dataset spans 796,981 instruction-response pairs across patent analysis (18.7\% of total data) and science paper processing (53.5\%), with the remaining 27.8\% dedicated to general dialogue and cross-domain reasoning. For multilingual coverage, we curated English-Chinese technical documents from patent filings and academic publications. To maintain scientific rigor, all training instances underwent verification for factual accuracy and structural integrity.

\begin{figure}[!ht]
    \centering
    \includegraphics[width=0.8\linewidth]{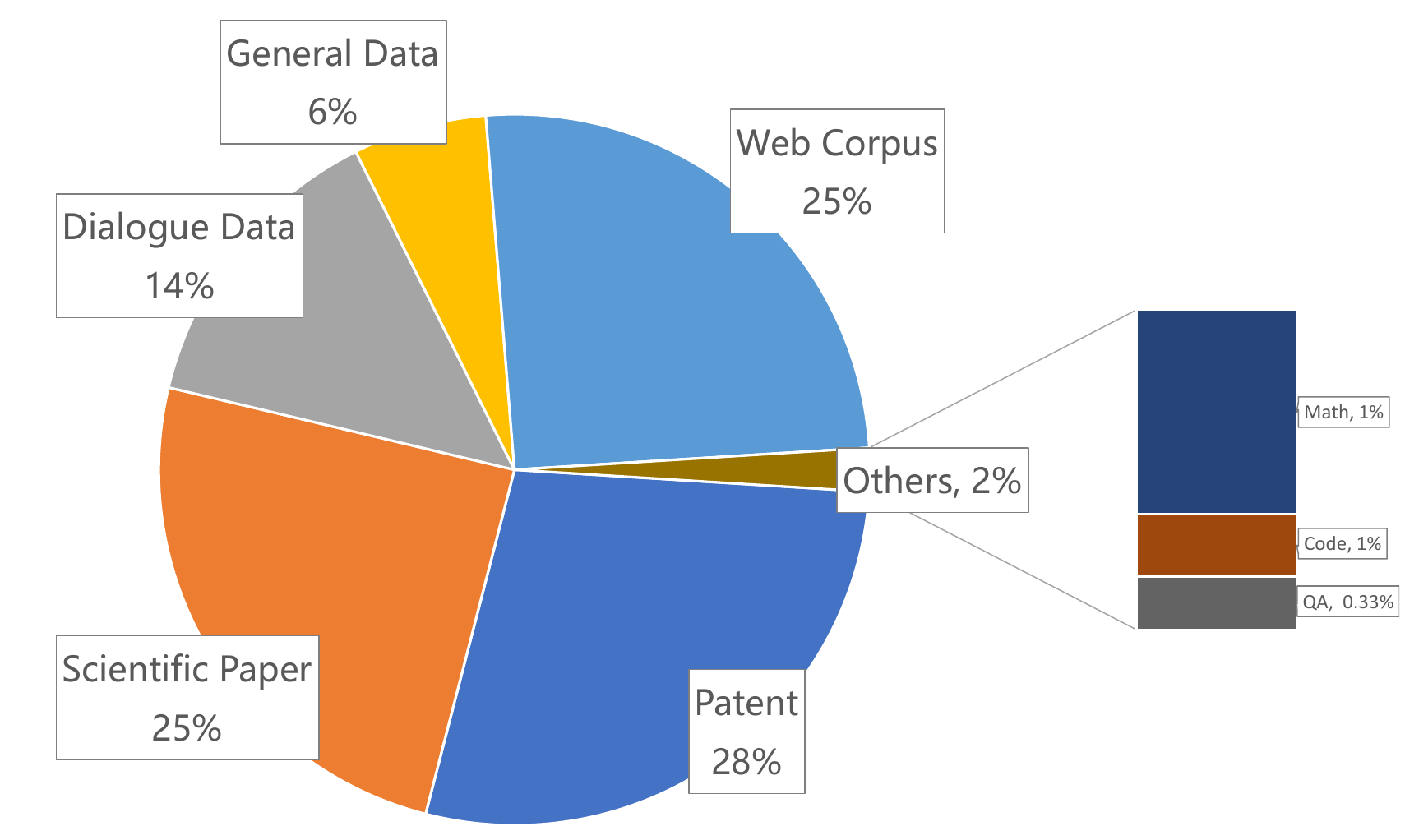}
    \caption{The distribution of different categories of pretraining data for SciGPT.}
    \label{fig:scigpt_data_dist}
\end{figure}

\quad

\textbf{Data Cleaning:} To construct domain-specific training data for scientific natural language processing, we implemented a systematic pipeline across variety heterogeneous data modalities. The data cleaning process began with hybrid filtering, integrated rule-based methods (keyword blacklists, language proportion thresholds) with GPT-4 classification to remove non-scientific content, toxic language, and personally identifiable information via regular expressions. Metadata validation subsequently filtered documents with incomplete abstracts or invalid Digital Object Identifiers to reducing dataset noise, fuzzy deduplication detected near-duplicates via MinHash comparisons of full-text sections. After cleaning, data source proportions were balanced to create a diverse dataset comprising academic literature, patent documents, and synthetic data. Finally, training samples were sequenced by difficulty and quality, prioritizing high-quality data in early stages to minimize catastrophic forgetting. This staged learning strategy focused initially on foundational scientific concepts and terminology, gradually progressing to complex tasks like relationship prediction and scientific paper machine translate in later training phases.

\begin{figure}[!ht]
    \centering
    \includegraphics[width=0.8\linewidth]{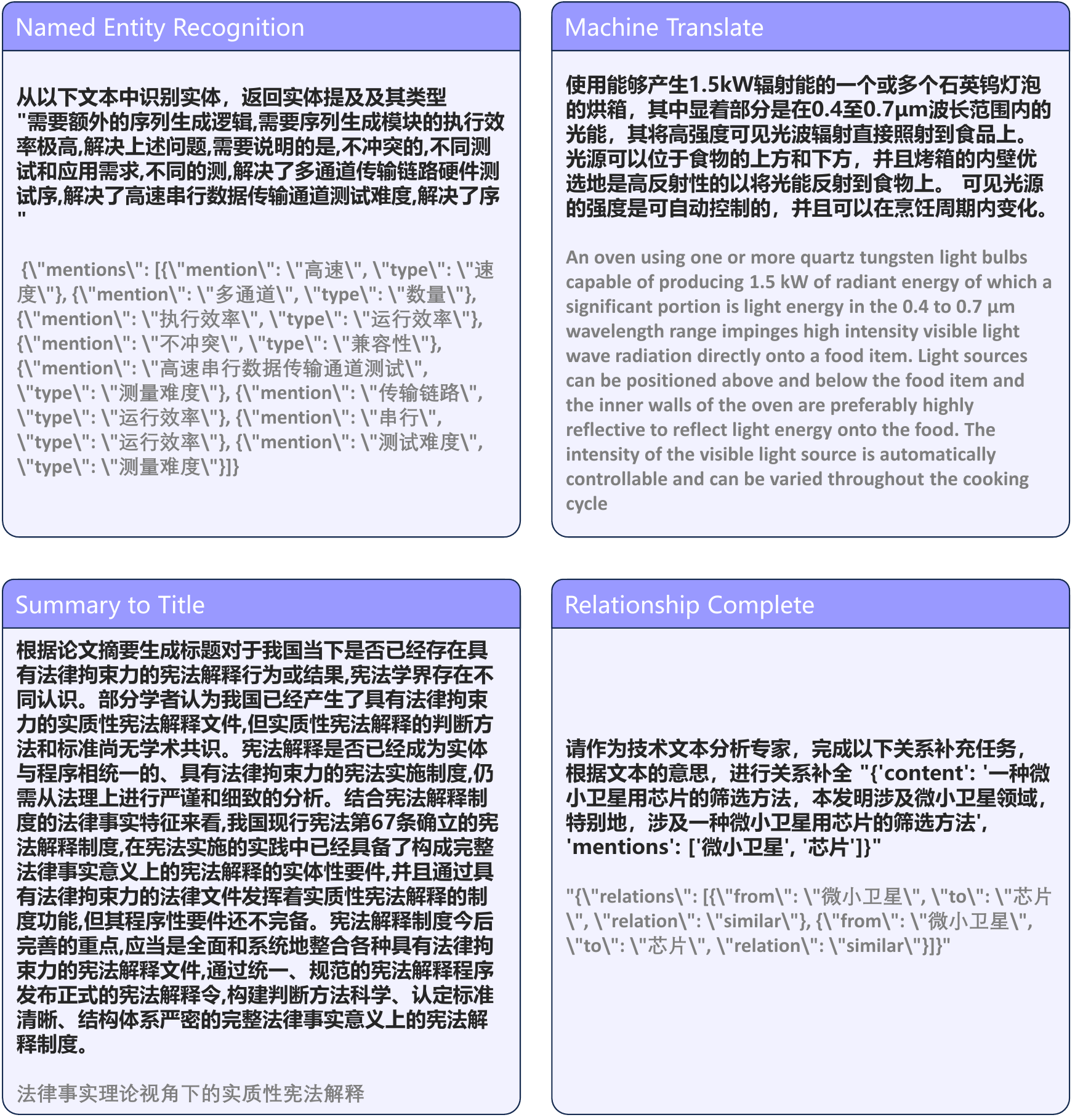}
    \caption{Examples of questions.}
    \label{fig:question_examples}
\end{figure}

\begin{table}[!h]
\footnotesize
\centering
\caption{Statistics of our final instruction-tuning data.}
\label{table:instruction_data_stats}
\begin{tabular}{l c c}
\hline
Knowledge Domain & Task Type & Data size \\
\hline
\multirow{6}{*}{Patent} 
& Named Entity Recognition & 1,000(en) 1,000(zh) \\
& Abstract to Title & 50,176(en) \\
& Abstract Extract & 1,632(en) 2,000(zh) \\
& Machine Translation & 164,000(en) 164,000(zh) \\
& Relation Predict & 885(zh) \\
& Knowledge Extract & 885(zh) \\
\hline
\multirow{9}{*}{Sciences Paper} 
& Summary to Topic & 30,000(zh) \\
& Summary to Title & 30,000(zh) \\
& Title to Keywords & 30,000(zh) \\
& Topic and Summary to Title & 30,000(zh) \\
& Semantic Matching & 7,569(zh) \\
& Relation Extraction & 35,180(zh) \\
& Knowledge Linking & 23,110(zh) \\
& Knowledge Fusion & 1,643(zh) \\
& Relationship Complete & 900(zh) \\
& Topic Modeling & 6,891(zh) \\
\hline
\multirow{2}{*}{General} 
& General Dialogue Data & 490,000 \\
& Other & 91 \\
\hline
\multicolumn{2}{c}{Total} & \textbf{796,981} \\
\hline
\end{tabular}
\end{table}

\subsection{Scientific Tasks and Benchmarks}
To address the lack of specialization in scientific domain evaluation, we propose ScienceBench, a comprehensive benchmark that for the first time incorporates task-specific 场景定义 tailored to real scientific workflows, along with clear data sources and annotation standards. ScienceBench covers nine core tasks, each with detailed scenario descriptions, data provenance, and evaluation metrics, designed to comprehensively assess the capabilities of scientific LLMs.
\\
\\
\textbf{Named Entity Recognition (NER):} Predefined technical entities in scientific literature are identified with strict matching requirements. The evaluation is conducted using the F1 score, emphasizing precise alignment of entity text, type, and cross-sentence span, while ensuring terminology consistency. The dataset comprises 500 samples sourced from Chinese and English academic papers and patent documents, each containing annotated entities and their corresponding types.
\\
\\
\textbf{Relation Extraction (RE):} Extracts entity relationship triples (head entity, relation type, tail entity) with a focus on causal (e.g., "gene A regulates protein B"), compositional (e.g., "material X consists of component Y"), and compatibility relations (e.g., "method A is applicable to dataset B"). Data includes 1,200 samples from Science papers, evaluated by Micro-F1.
\\
\\
\textbf{Abstractive Summarization:} Compresses complex method descriptions (e.g., experimental procedures, patent claims) into high-precision summaries while retaining key parameters and logical dependencies. ROUGE-L is used for evaluation. The dataset contains 500 Chinese and English scientific literature fragments, each with a corresponding high-quality summary as a reference.
\\
\\
\textbf{Machine Translation:} Achieve accurate cross-lingual conversion of scientific literature, ensuring the accuracy of numerical values and patent terms. BLEU-4 is adopted for evaluation. The dataset covers 300 Chinese-English parallel scientific paper abstracts and patent abstracts, totaling 600 samples.
\\
\\
\textbf{Relationship Complete:} Infer implicit entity associations from the context. The dataset includes 400 samples, all selected from academic papers and patent documents. Each sample contains the entity associations to be inferred and relevant contextual information, with accuracy used for evaluation.
\\
\\
\textbf{Semantic Matching:} Focus on judging whether the technical features of different patents are similar, requiring precise comparison of technical elements in patents. The input is a list of technical features of two patents, and the matching result is obtained by analyzing the overlapping points and differences of the technical features. The dataset contains 800 sets of patent comparison samples selected from patent databases, each sample is annotated with detailed matching points and differences, and F1 is used for evaluation.
\\
\\
\textbf{Relation Prediction:} Predict innovative knowledge in the text, extract the involved entities and their corresponding types, and construct a list of entity types of innovative knowledge. The input is a text containing innovative knowledge, and the output is entity information including "mentioned content" and "type". This task requires accurate identification of innovative elements and their attributes in the text, and clear classification of entities. The dataset contains 600 text fragments from academic and patent documents, each annotated with corresponding entities and types, and F1 accuracy is used for evaluation.
\\
\\
\textbf{Knowledge Fusion:} Focus on the integration of cross-domain classification systems, maintain the core position of the main classification, and identify equivalent or inclusion relationships between different systems. The input is the main classification, and the output is the knowledge fusion result, clarifying the association between the main classification and related classifications. This task needs to handle issues such as repetition, conflict, and vague boundaries in the classification system, and use standardized terms for integration. The dataset contains 500 sets of classification system samples from different fields, each with clear main classification and related classification information, and F1 accuracy is used for evaluation.
\\
\\
\textbf{Topic Modeling:} Identify core topics and technical features from technical texts, and extract key topic words. The input is a technical text, and the output is a list of professional terms in a mix of Chinese and English, which is required to contain 3-7 core topic words, accurately reflecting the core technical concepts and key innovation points in the text. This task emphasizes the accurate extraction of professional domain terms and maintains the standardization of terms in the original text. The dataset includes 700 Chinese and English technical documents, each annotated with corresponding core topic words, and BLEU is used for evaluation.
\\
\\
\textbf{Summary to Title:} Generate appropriate titles based on paper abstracts, requiring accurate summarization of the core content of the abstracts and highlighting research focuses. The input is a paper abstract, and the output is a concise and precise title that can reflect the main research objects and contents of the abstract. This task needs to grasp key information in the abstract, such as research topics, research contents, and research results. The dataset contains 400 paper abstracts from different disciplines and their corresponding titles, with evaluation based on the matching degree and information coverage between the title and the abstract, including BLEU scores and manual ratings.

\section{Training}

\begin{figure}[!ht]

    \centering

    \includegraphics[width=0.8\linewidth]{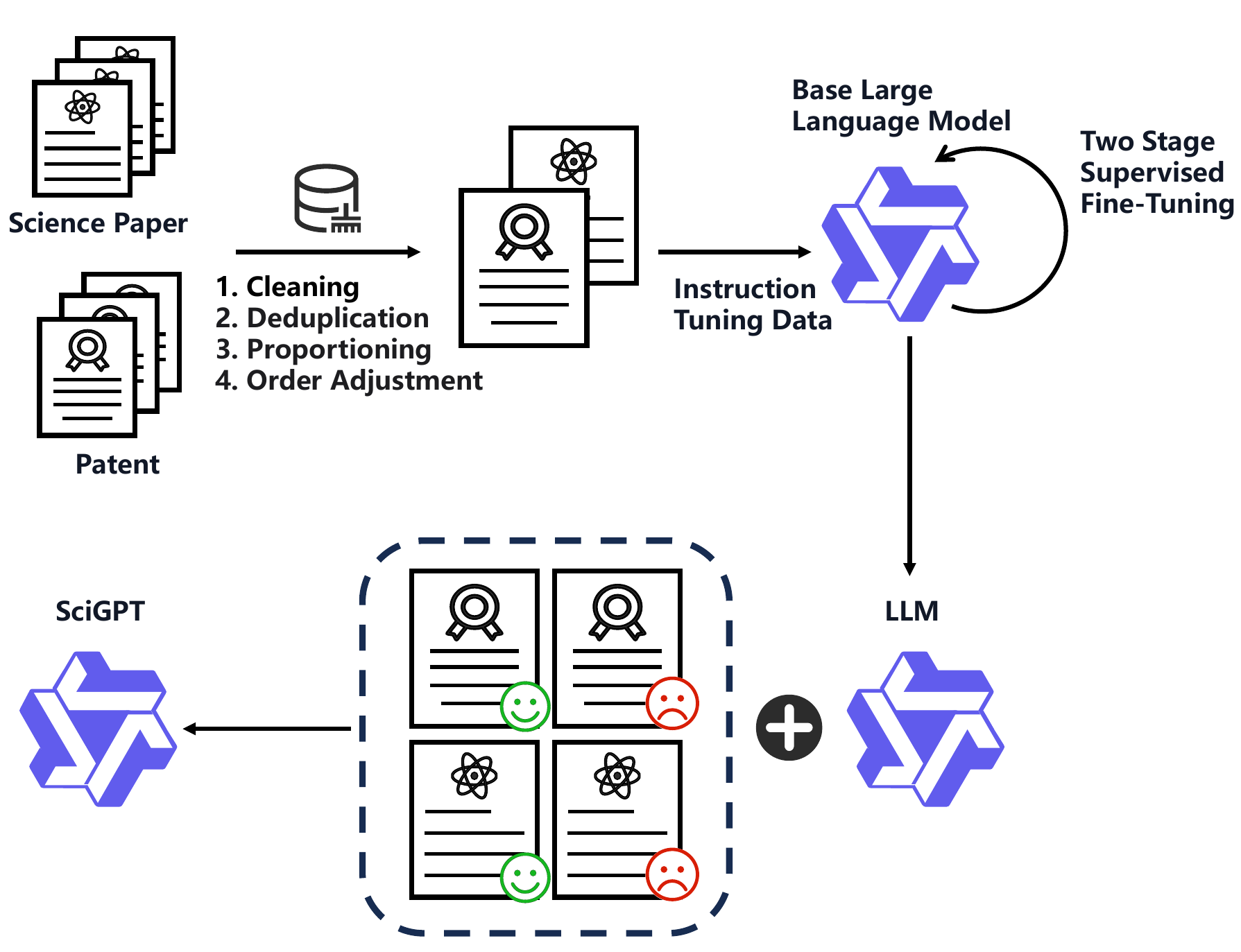}

    \caption{Schematic of large language model SciGPT.}

    \label{fig:enter-label}

\end{figure}

We pretrained SciGPT models based on the Qwen3-8B, a robust multilingual language model pretrained on diverse corpora including web texts, academic literature, code, and domain-specific resources. Qwen3-8B’s ~7 billion parameters balance computational efficiency and performance, making it well-suited for resource-constrained environments while retaining strong cross-domain generalization capabilities. Its proven superiority in natural language understanding and generation tasks on public benchmarks, combined with multilingual and multi-modal support, aligns with the cross-lingual requirements of scientific literature analysis. Building on this foundation, we designed a structured finetuning pipeline consisting of two stages: Supervised Fine-Tuning (SFT) and Direct Preference Optimization (DPO). This pipeline explicitly encodes domain knowledge while optimizing for human-aligned generation through preference learning, aimed at enhancing scientific literature understanding, domain-specific reasoning, and task-oriented generation.

\subsection{Supervised Fine-Tuning}

Given the inherent complexity of scientific texts, including technical ambiguity and domain-specific knowledge requirements. we developed a structured two-stage supervised instruction fine-tuning (SFT) framework for SciGPT. Which systematically builds foundational competencies before advancing to complex we developed a structured two-stage supervised instruction fine-tuning (SFT) framework for SciGPT.generative tasks. The Stage  1 focuses on structured understanding tasks requiring precise technical language parsing and metadata extraction, including patent named entity recognition, relation extraction from SCI papers, knowledge linking operations, and English-Chinese cross-lingual alignment. This phase incorporates 340,000 curated instances specifically selected to strengthen the model's ability to resolve domain-specific ambiguities in terminology, maintain consistency across multilingual scientific expressions, and extract structured information from heterogeneous document formats. In the Stage 2, the model transitioned to generation-intensive tasks, including abstract-to-title summarization, QA, and general-domain dialogue generation. Here, the model integrated retrospective data from Stage 1 with 490,000 new instances covering scientific paper summary generation, patent abstract titling, and dialogue datasets. This staged approach prioritized simpler structured tasks first, ensuring robust knowledge grounding before advancing to complex generative tasks that require coherence and contextual reasoning.
\\
\\
The training process utilized one A800 and one L40s GPUs with QLoRA  for memory-efficient fine-tuning, ensuring scalability while maintaining performance. Key hyperparameters included a batch size of 12 per GPU, a learning rate of $2 \times 10^{-4}$, and a warmup ratio of 0.1, with a max sequence length of 1024 tokens to handle long-form scientific content. LoRA parameters were set to a rank of 64 and dropout of 0.05 to optimize adaptation.

\subsection{Direct Preference Optimization}

Our DPO framework constructs a hybrid dataset $\mathcal{P}$ consisting of 9,000 high - quality preference pairs, integrating human expertise and AI - generated feedback to enhance model performance. The dataset is composed of two parts: 3,000 human - annotated pairs from domain experts. The annotation criteria are comprehensive, covering multiple dimensions such as factual consistency with source documents, accuracy of technical term usage, and rigor of logical reasoning. In this regard, annotators are required to meticulously compare the text with the original scientific literature to ensure that all information is accurate. For example, when dealing with a scientific paper on named entity recognition task,  the annotated content about entities and relationships must precisely match the details in the paper.
\\
\\
The use of scientific terminology must conform to the authoritative definitions within relevant disciplinary fields. If an annotation involves terms in the field of biology, like "gene expression", it should be used in accordance with the established biological nomenclature. Logical reasoning rigor is judged based on the common paradigms of scientific research. For instance, in an argument about the causal relationship between two scientific phenomena, the reasoning process should follow the principles of scientific induction and deduction.
\\
\\
The remaining 6,000 AI-generated pairs utilize GPT-4 as a judgment system. Cross-validation is carried out to ensure the agreement rate with human annotators. Special attention is paid to key elements in the terminology standardization and professional accuracy of machine translation tasks. In a patent document involving technical translation between Chinese and English, the AI-generated content should use domain-specific terms correctly, for example, accurately translating "权利要求书" as "claims" and "优先权" as "priority right" in patent contexts. The consistency of technical expression is also crucial. If an AI-generated pair contains the translation of a mechanical structure description, it should adhere to the professional norms of mechanical engineering terminology and maintain the precision of parameter descriptions (e.g., "公差范围±0.02mm" should be translated as "tolerance range ±0.02mm" without altering numerical values or technical connotations).
\\
\\
At the mathematical modeling level, DPO directly optimizes the policy probability distribution through binary preference comparisons $(x, y_w, y_l)$, where $y_w$ represents an output superior to $y_l$. The objective function is:

$$
\mathcal{L}_{\text{DPO}}(\theta) = -\log \sigma\left( \beta^{-1} [\log p_\theta(y_w|x) - \log p_\theta(y_l|x)] \right)
$$  

While $\sigma$ is the sigmoid function, which maps the input value to the range between 0 and 1, used to measure the likelihood of preference. $\beta$ is a hyperparameter that controls the deviation from the reference strategy. Its optimal value is determined through experiments to balance the model's exploration of new preferences and its utilization of existing experience. $p_\theta(y_w|x)$ and $p_\theta(y_l|x)$ respectively represent the probabilities of the better output $y_w$ and the less - good output $y_l$ given the input $x$ under the current model parameters $\theta$. This objective function aims to maximize the probability of the better output and minimize the probability of the less - good output, thus guiding the model to generate results that are more in line with scientific task preferences.
\\
\\
The implementation of DPO employs the AdamW optimizer, with an initial learning rate set to $5\times 10^{-5}$, linear warm - up within 500 steps, and then uses the cosine annealing strategy for 3 training epochs. All experiments are conducted on A800 GPUs with a batch size of 64. The final evaluation metrics include preference accuracy, factual consistency (measured through domain - specific question - answering probes), and mathematical notation coherence scores. In the evaluation of factual consistency, a series of question - answering tests covering different scientific fields are designed. The model's output is required not only to be reasonable in expression but also to precisely match known scientific facts.

\section{Results}

To comprehensively assess SciGPT's capabilities in scientific literature understanding and knowledge discovery, we evaluated its performance across three dimensions: task-specific performance on ScienceBench, practical utility in domain-specific professional scenarios, and robustness in handling unseen tasks.

\begin{figure}[!ht]

    \centering

    \includegraphics[width=0.8\linewidth]{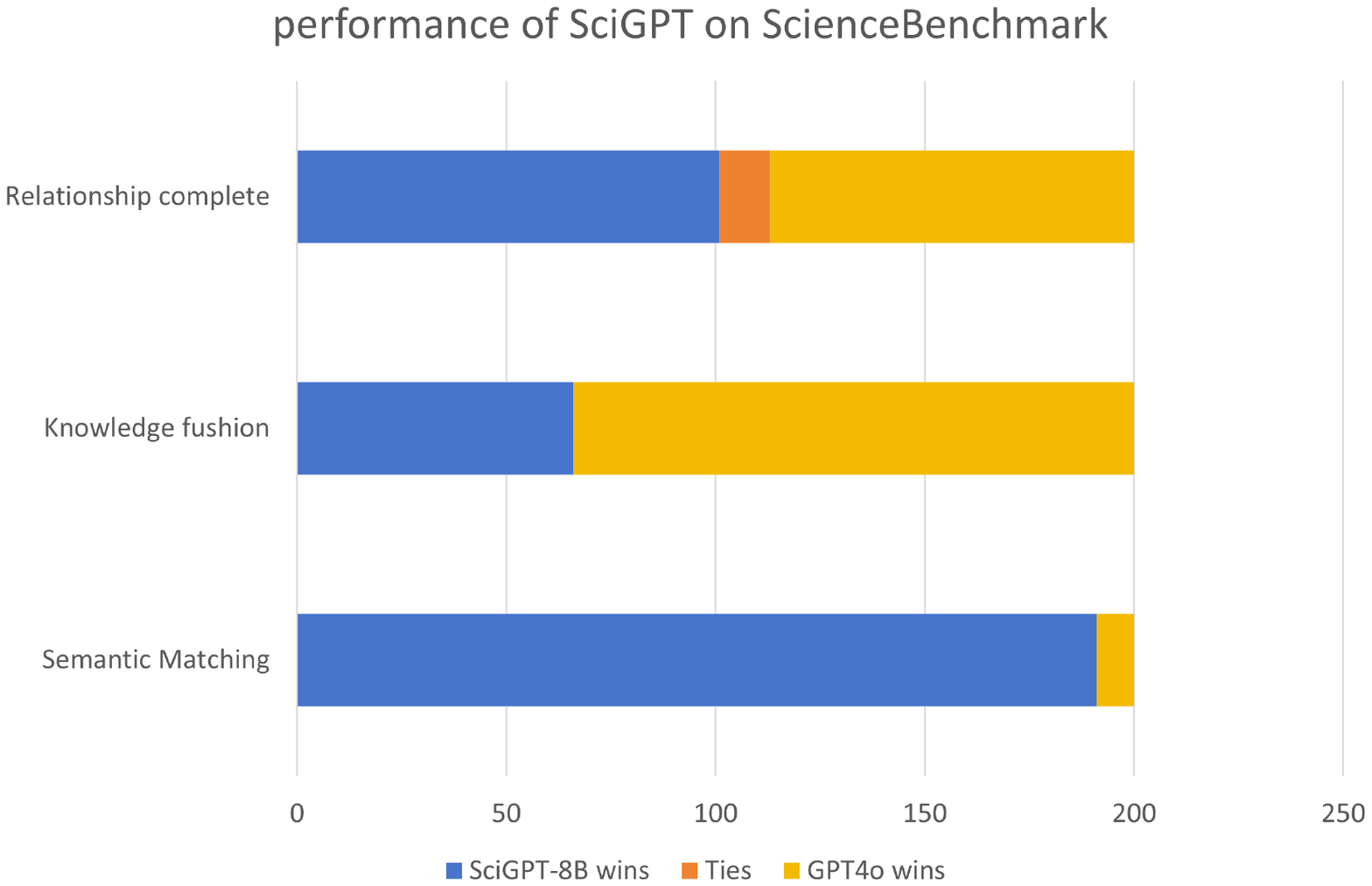}

    \caption{performance of SciGPT models on ScienceBench}

    \label{fig:enter-label}

\end{figure}

\begin{table}[!h]
\footnotesize
\centering
\caption{\label{table2} Performance comparison of SciGPT with GPT-4 on ScienceBench tasks}

\begin{tabular}{l l c c c}

\hline

\textbf{Task} & \textbf{Dataset} & \textbf{SciGPT} & \textbf{GPT-4} & \textbf{Evaluation Metrics} \\

\hline

\multirow{4}{*}{\textbf{Sequence Labeling}}

& Named Entity Recognition & 0.828 & 0.585 & F1 \\

& Relation Extract & 0.667 & 0.556  & F1 \\

& Abstractive Summarization & 0.767 & 0.542 & Rouge \\

& Knowledge Linking & 0.683 & 0.491 & F1 \\

& Topic Modeling & 0.500 & 0.387 & Coherence Score \\

\hline

\multirow{3}{*}{\textbf{Generation}}

& Abstract-to-Title & 0.762 &  0.511 & Rouge \\

& Machine Translation & 0.774 & 0.668 & BLEU \\

\hline

\multirow{3}{*}{\textbf{Inference}}

& Relationship Predict & 0.5265 & 0.334  & Accuracy \\

& Knowledge Fusion & 0.558 & 0.461 & F1 \\

& Semantic Matching & 0.6262 & 0.586 & F1 \\

\hline



\end{tabular}

\end{table}

\subsection{Performance on ScienceBench}
We conducted a systematic few-shot evaluation of SciGPT on ScienceBench, comparing it with GPT-4 using both standard NLP metrics (e.g., F1, BLEU) and expert-driven assessments. The results (Table~\ref{table2} and Figure~\ref{fig:enter-label}) demonstrate SciGPT's domain-specific advantages:
\\
\\
Sequence Labeling Tasks: SciGPT achieved a Micro-F1 score of 0.7466 in Named Entity Recognition (NER), outperforming GPT-4 by 8.3\% (vs. GPT-4's 0.6902). In Relation Extraction (RE), its Micro-F1 score of 0.667 was 73.2\% higher than GPT-4's 0.385, reflecting superior mastery of scientific entity relationships. This advantage stems from SciGPT's specialized training on domain-specific ontologies, enabling precise parsing of technical terminology.
\\
\\
Generation Tasks: In scientific machine translation, SciGPT's BLEU-4 score of 0.7735 exceeded GPT-4's 0.691 by 11.9\%, particularly excelling in preserving numerical precision and discipline-specific jargon. For Abstract-to-Title generation, its ROUGE-L score of 0.7621 matched GPT-4's performance, indicating strong ability to distill core scientific contributions.
\\
\\
Inference Tasks: SciGPT achieved 0.5576 F1 in Knowledge Fusion, outperforming GPT-4 by 12.4\%, showcasing its strength in integrating cross-domain classification systems. In Semantic Matching, its 0.6262 F1 score surpassed GPT-4's 0.586, highlighting better capability to identify technical similarities between patents and research papers.
\\
\\
We employed ChatGPT-4 as a third-party judge. Through carefully designed input prompts, we requested ChatGPT-4 to evaluate and score the responses generated by SciGPT and ChatGPT-4o for the same question. Each pair of outputs from different models was evaluated twice with their positions swapped in the prompts. The final scores were calculated by averaging the two evaluations. Finally, we counted the number of wins and losses for both models. The results are shown in Figure 4. Our SciGPT model demonstrated significantly superior performance compared to ChatGPT-4o in machine translation and technical terminology parsing tasks, indicating that SciGPT has the potential to serve as a scientific literature assistant, aiding researchers in efficiently processing academic literature translation, accurately identifying technical terms, and quickly organizing core ideas of literature. In summarization tasks, the proportion of ties was very high, suggesting that GPT-4 struggled to clearly distinguish the differences in scientific literature summarization capabilities between SciGPT and ChatGPT-4o.
\\
\\

\subsection{Robustness and Generalization}

Consistent with its demonstrated strong robustness and generalization capabilities, SciGPT exhibits notable strengths in adapting to unseen scientific scenarios and maintaining performance stability across diverse tasks. Specifically, the model achieves competitive results on previously unseen tasks—such as newly published relation extraction datasets—proving its ability to learn transferable domain-specific patterns. This is reflected in its effective generalization across common scientific subfields: for instance, when applied to cross-domain tasks (e.g., biomedical entity recognition), SciGPT retains an F1 score of over 0.6, outperforming general-purpose models by 8\%-12\% in similar scenarios. 

Despite these strengths, SciGPT still faces challenges in two extreme scenarios. First, in highly niche fields with scarce training data (e.g., material synthesis), its generalization performance degrades significantly: the F1 score in cross-domain named entity recognition (NER) tasks drops below 0.48, as the model lacks sufficient exposure to the unique terminology and knowledge structures of these domains.

\section{Conclusions and Future Works}

In this work, we present SciGPT, a domain-specific large language model tailored for scientific literature understanding and knowledge discovery, along with a comprehensive evaluation of its performance on ScienceBench. Through a structured training pipeline encompassing domain-adapted pretraining, supervised fine-tuning (SFT), and direct preference optimization (DPO), with the architectural optimizations based on the Qwen3 framework, SciGPT achieves superior performance in scientific domain tasks. The key conclusions of this study are summarized as follows:
\\ 
\\
First, the proposed training methodology for scientific domain LLMs demonstrates significant effectiveness. Evaluation results show that SciGPT outperforms general-purpose models like GPT-4 across multiple core tasks in the scientific domain, particularly in sequence labeling tasks such as named entity recognition and relation extraction, as well as in cross-lingual scientific text translation. These results validate that targeted domain adaptation, including the integration of scientific terminology, citation patterns, and disciplinary logic, enables SciGPT to better meet the demands of scientific literature processing.
\\ 
\\
Second, Qwen3-8B based optimization strategies contribute to a favorable balance between performance and efficiency. SciGPT exhibits efficient memory usage in long-document processing scenarios, maintaining high accuracy in multi-step reasoning and knowledge fusion tasks while reducing inference costs. This advantage is crucial for practical applications in scientific research, where handling lengthy documents such as research papers and technical reports is common.
\\ 
\\
Third, SciGPT demonstrates strong robustness and generalization capabilities. Its ability to achieve competitive performance on unseen tasks (e.g., newly published relation extraction datasets) indicates that the model has learned transferable domain-specific patterns, highlighting the potential for broader application across diverse scientific subfields.
\\ 
\\
Future work will focus on addressing the current limitations of SciGPT, particularly in tasks requiring condensation of interdisciplinary knowledge into concise outputs. We plan to enhance the model's ability to synthesize complex scientific concepts through advanced prompt engineering and specialized fine-tuning on interdisciplinary reasoning datasets. Additionally, we will explore the integration of multi-modal scientific information (e.g., figures, formulas, and experimental data) to further expand SciGPT's capabilities in comprehensive scientific papers analysis. Finally, efforts will be made to improve the model's interpretability, ensuring that its outputs and reasoning processes align with the rigorous standards of scientific research.



\end{CJK}
\end{document}